\documentclass[11pt,a4paper]{article}

\usepackage{acet}
\usepackage{times}
\usepackage{latexsym}
\usepackage{url}

\usepackage{color,graphicx} 

\title{Evaluating the Impact of Khmer Font Types on Text Recognition}

\author{
  Vannkinh Nom$^{\dag\ddag}$ \quad
  Souhail Bakkali$^{\dag}$ \quad
  Muhammad Muzzamil Luqman$^{\dag}$ \\
  Mickaël Coustaty$^{\dag}$ \quad
  Jean-Marc Ogier$^{\dag}$ \\
  \\
  $^\dag$La Rochelle University, Laboratoire Informatique Image Interaction (L3i) \\
  $^\ddag$Cambodia Academy of Digital Technology (CADT) \\
  \texttt{\{vannkinh.nom, souhail.bakkali, muhammad\_muzzamil.luqman,} \\
  \texttt{mickael.coustaty, jean-marc.ogier\}@univ-lr.fr}
}

\summary{
Text recognition is significantly influenced by font types, especially for complex scripts like Khmer. The variety of Khmer fonts, each with its unique character structure, presents challenges for optical character recognition (OCR) systems. In this study, we evaluate the impact of 19 randomly selected Khmer font types on text recognition accuracy using Pytesseract. The fonts include Angkor, Battambang, Bayon, Bokor, Chenla, Dangrek, Freehand, Kh Kompong Chhnang, Kh SN Kampongsom, Khmer, Khmer CN Stueng Songke, Khmer Savuth Pen, Metal, Moul, Odor MeanChey, Preah Vihear, Siemreap, Sithi Manuss, and iSeth First. Our comparison of OCR performance across these fonts reveals that Khmer, Odor MeanChey, Siemreap, Sithi Manuss, and Battambang achieve high accuracy, while iSeth First, Bayon, and Dangrek perform poorly. This study underscores the critical importance of font selection in optimizing Khmer text recognition and provides valuable insights for developing more robust OCR systems.
}

\begin{document}
\maketitle

\medskip
\noindent{\bf Keywords:} Text Recognition, Khmer Script, Pytesseract

\section{Introduction}

Text recognition technology has revolutionized the way we interact with written content, enabling machines to understand and process textual information from various sources. However, the accuracy of Optical Character Recognition (OCR) systems can vary significantly based on several factors, including the font types used in the text. This variability is particularly pronounced in scripts with rich typographic diversity, such as the Khmer script. The Khmer language, spoken by the majority of Cambodians, features a horizontal writing system that progresses from left to right and downwards ~\cite{born2022encoder}. The script encompasses a broad range of font styles, each with distinct characteristics and visual elements, which can introduce unique challenges for OCR systems. Depending on the various Khmer fonts, some groups of characters may appear similar in shape but are actually distinct, while others may look different depending on the style and font type.

\begin{figure}[tp]
\begin{center}
\includegraphics[width=7cm]{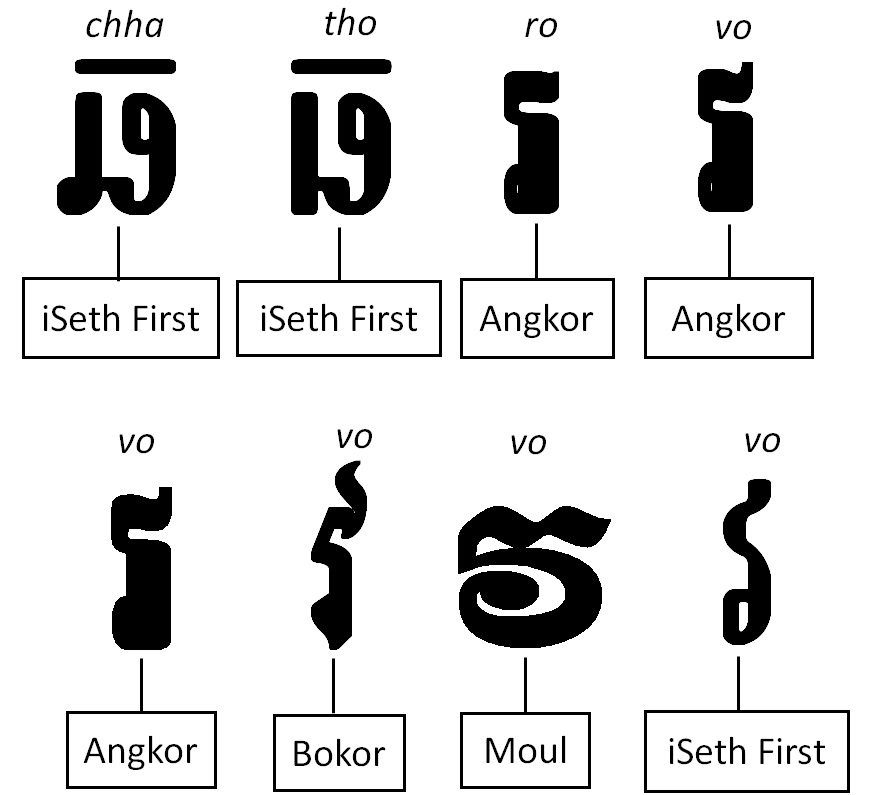}
\caption{Examples of how characters can be displayed: In the first row, the characters are distinct but may appear similar due to being in the same font. In the second row, the same character is shown in different shapes across various font types.} \label{sampleimg}
\end{center}
\end{figure} 

In addition to font variability, the complexity of Khmer script itself presents challenges for OCR accuracy. The script includes 33 consonants, 16 dependent vowels, 14 independent vowels, and 13 diacritics ~\cite{horton2017spoof}. According to the Guinness World Records, Khmer script holds the record for the longest alphabet, consisting of 74 distinct characters ~\cite{valy2019text}. Unlike Latin scripts, Khmer words are constructed in a unique manner. Consonants can take different shapes based on their position within a word, featuring both primary and secondary consonants. Furthermore, consonants can merge into new forms, known as subscripts or low consonants, which appear below the main consonant  ~\cite{valy2017new}. These complexities in character formation, along with the diversity of fonts, make Khmer script particularly challenging for OCR systems to process accurately.
Figure 1 illustrates the complexity of Khmer script by highlighting examples of characters that, while visually similar, represent entirely different characters depending on the font style. It also shows characters with similar shapes but distinct meanings, even when they are in the same font. As Khmer text recognition becomes increasingly vital in digital applications, understanding how different font types and the script's inherent complexity affect recognition accuracy is crucial for optimizing OCR performance.

Our research involves a comprehensive analysis of 19 Khmer font types, encompassing various writing styles. Each font is evaluated for its impact on text recognition accuracy, measured by the character error rate (CER) and word error rate (WER). This analysis identifies which fonts are most easily and most challenging to recognize using Pytesseract when processing Khmer script. The findings are intended to guide the selection of optimal fonts for text recognition applications and to contribute to the broader field of OCR research by addressing the specific challenges posed by Khmer font diversity and script complexity. As OCR technology continues to evolve, understanding the nuances of font impact and script intricacies on recognition accuracy will be essential for advancing text recognition systems in multilingual and diverse script environments.

\section{Related Work}
Optical Character Recognition (OCR) is a widely researched field with extensive academic and practical interest, particularly for its role in automating text extraction from various sources. Pytesseract, a popular Python tool, has been widely applied in this domain, serving diverse purposes. 
A recent study ~\cite{dome2021optical} proposes leveraging Pytesseract, an OCR tool, in conjunction with classification techniques to enhance the accuracy and efficiency of text extraction. By integrating Pytesseract with classification methods, the research aims to improve the automatic extraction of specific information from paper documents, potentially reducing or even eliminating the need for costly data entry work. 
The study conducted by ~\cite{soumya2024handwritten} suggests utilizing Pytesseract for recognizing handwritten text, preceded by a three-step preprocessing procedure. First, the images are converted to grayscale to minimize contrast. Next, Gaussian blur is applied to reduce noise and enhance the image. Lastly, thresholding is implemented to convert the image into a binary format, improving the separation between the text and the background for clearer recognition.
The research by ~\cite{bannigidad2023ancient} utilized the Pytesseract-OCR engine to recognize characters in Ancient Kannada texts, leveraging it for optical character recognition (OCR) purposes. This engine, a Python wrapper for the Tesseract OCR tool, is capable of reading text from images. The researchers improved image quality through binarization, followed by segmentation using the blob technique. They then used regional zoning to achieve character-level segmentation.
The research carried out by ~\cite{mahajan2022natural} proposed combining the Efficient and Accurate Scene Text Detector (EAST) with Pytesseract for detecting and recognizing text in natural scenes. EAST is used to pinpoint areas within images that are most likely to contain text, while Pytesseract extracts and analyzes this text.
Additionally, ~\cite{chadha2020license} developed a License Plate Recognition System using Python, OpenCV, and Pytesseract, achieving reliable results even in challenging environments. The process begins by converting the vehicle images to grayscale, detecting closed shapes that could represent characters on the license plates, and then applying OCR to extract the text from the grayscale images.

Given the robust capabilities of Pytesseract, we have chosen to utilize this tool in our research on text recognition across various Khmer script font types to evaluate how different fonts impact text recognition accuracy.

\section{Dataset}

\begin{table}[!ht]
\caption{Examples of different types of Khmer fonts display various styles.}
\label{fontstyle}
\vspace{0.3cm}
\centering
\resizebox{1\columnwidth}{!}{%
\begin{tabular}{llcc} \hline\hline
No & Khmer Font Types & Example \\ \hline
1 & Angkor & \includegraphics{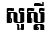} \\
2 & Battambang & \includegraphics{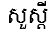} \\
3 & Bayon & \includegraphics{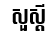} \\
4 & Bokor & \includegraphics{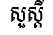} \\
5 & Chenla & \includegraphics{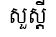} \\
6 & Dangrek & \includegraphics{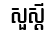} \\
7 & Freehand & \includegraphics{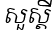} \\
8 & Kh Kompong Chhnang & \includegraphics{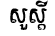} \\
9 & Kh SN Kampongsom & \includegraphics{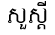} \\
10 & Khmer & \includegraphics{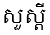} \\
11 & Khmer CN Stueng Songke & \includegraphics{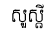} \\
12 & Khmer Savuth Pen & \includegraphics{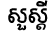} \\
13 & Metal & \includegraphics{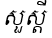} \\
14 & Moul & \includegraphics{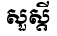} \\
15 & Odor MeanChey & \includegraphics{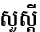} \\
16 & Preah Vihear & \includegraphics{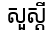} \\
17 & Siemreap & \includegraphics{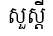} \\
18 & Sithi Manuss & \includegraphics{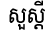} \\
19 & iSeth First & \includegraphics{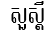} \\
\hline\hline
\end{tabular}
}
\end{table}

Khmer text recognition presents unique challenges due to the vast number of font variations that exist within the script. Each font style brings its own structural characteristics, making the process of accurate recognition more complex. To tackle this issue, we propose creating a synthetic dataset composed of black-and-white images that feature a wide variety of Khmer fonts. This synthetic dataset is designed to capture the diverse visual representations of the Khmer script, providing a robust foundation for improving text recognition algorithms. In our study, we carefully selected 19 different Khmer font types, ensuring a broad representation of popular and traditional fonts. These fonts include Angkor, Battambang, Bayon, Bokor, Chenla, Dangrek, Freehand, Kh Kompong Chhnang, Kh SN Kampongsom, Khmer, Khmer CN Stueng Songke, Khmer Savuth Pen, Metal, Moul, Odor MeanChey, Preah Vihear, Siemreap, Sithi Manuss, and iSeth First. The selection of these fonts was intended to encompass both modern and classic styles, as well as those used in formal and informal contexts, to create a dataset that reflects real-world usage of Khmer text. For each image in the dataset, we generated around 100 words using these font types. In addition to the image, we provide an accompanying XML file that contains detailed information about the text present in the image and the corresponding font type. This pairing of images and XML metadata allows for a more comprehensive analysis of the font's impact on text recognition, making it easier to correlate OCR performance with specific font types. Table 1 provides a detailed overview of the different font types used in the dataset, illustrating how each font displays distinct stylistic elements.


\section{Text Recognition}
Text recognition is a crucial process that involves identifying and extracting textual information from various sources.

\begin{table}[!ht]
\caption{Performance of different Khmer font types: character error rate (CER) and word error rate (WER).}
\label{fontstyle}
\centering
\vspace{0.3cm}
\begin{tabular}{llcc} \hline\hline
No & Khmer Font Types & CER & WER \\ \hline
1 & Angkor & 0.11 & 0.45 \\
2 & Battambang & \textbf{0.01} & \textbf{0.07} \\
3 & Bayon & 0.21 & 0.77 \\
4 & Bokor & 0.06 & 0.26 \\
5 & Chenla & 0.04 & 0.19 \\
6 & Dangrek & 0.29 & 0.97 \\
7 & Freehand & 0.14 & 0.45 \\
8 & Kh Kompong Chhnang & 0.16 & 0.62 \\
9 & Kh SN Kampongsom & 0.06 & 0.26 \\
10 & Khmer & \textbf{0.01} & \textbf{0.08} \\
11 & Khmer CN Stueng Songke & 0.09 & 0.37 \\
12 & Khmer Savuth Pen & 0.14 & 0.56  \\
13 & Metal & 0.11 & 0.44 \\
14 & Moul & 0.13 & 0.50 \\
15 & Odor MeanChey & 0.02 & 0.08 \\
16 & Preah Vihear & 0.10 & 0.43 \\
17 & Siemreap & 0.02 & 0.10 \\
18 & Sithi Manuss & 0.02 & 0.08 \\
19 & iSeth First & 0.23 & 0.82 \\
\hline\hline
\end{tabular}
\end{table}

\begin{figure*}[!ht] 
    \centering
    \includegraphics[width=1\linewidth]{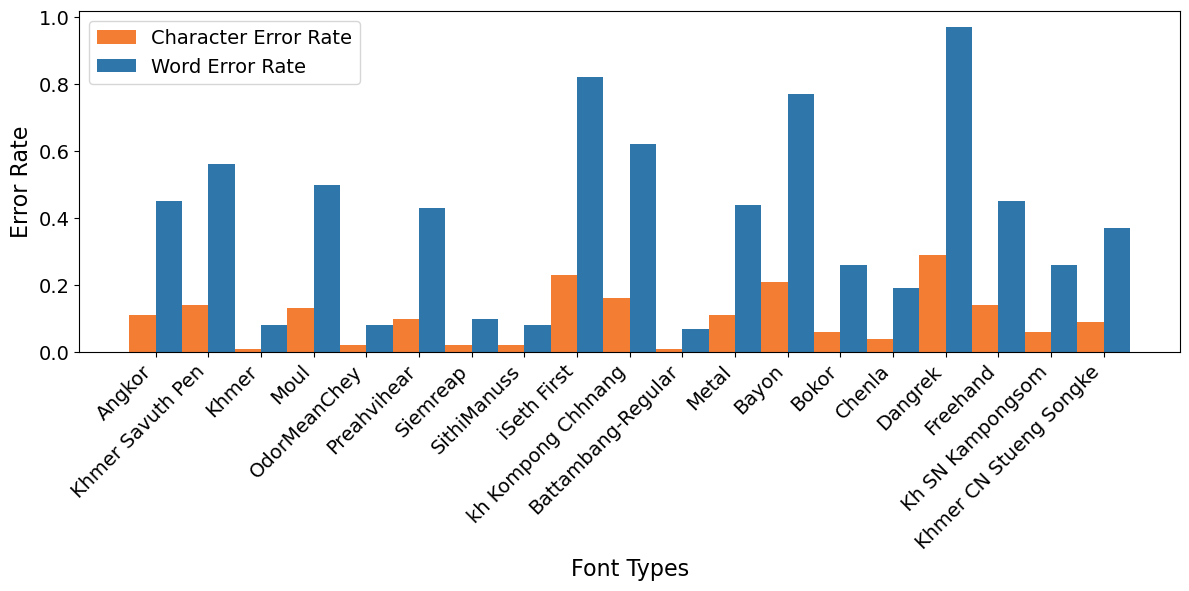}
    \caption{The character and word error rates for text recognition using Pytesseract show distinct results based on the font type.}
    \label{fig:1}
\end{figure*}

\begin{table*}[!ht]
  \caption{The most difficult characters to recognize are found in the Bayon, Dangrek, and iSeth First fonts.}
  \label{fontstyle}
  \vspace{0.2cm}
  \centering
  \begin{tabular}{lc} 
    \hline\hline
    Font & Characters \\ \hline
    \multicolumn{2}{c}{\includegraphics[width=0.7\textwidth]{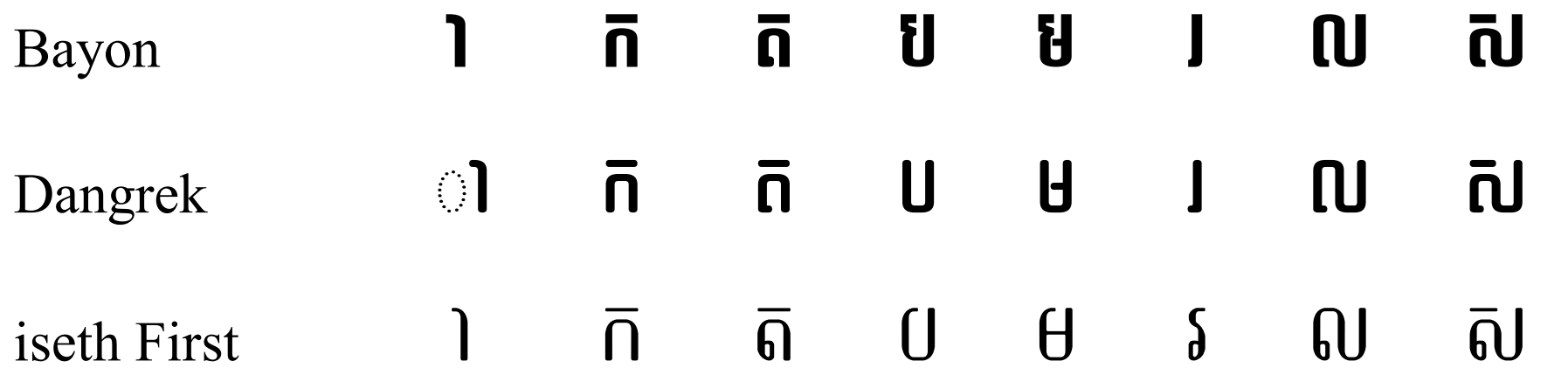}} \\ 
    \hline\hline
  \end{tabular}
\end{table*}

 We evaluated the performance of text recognition using Pytesseract, a popular optical character recognition (OCR) tool, on our specifically designed dataset. To assess the effectiveness of text recognition, we employed two key metrics: the character error rate (CER) and the word error rate (WER). For this purpose, we used Jiwer version 3.0.4, a Python library known for its efficiency in calculating these error rates. Our experimental setup involved using Pytesseract version 0.3.10 to extract text from images featuring a range of different font types. We carefully selected these fonts to ensure a comprehensive evaluation of the OCR tool's performance. By analyzing the accuracy of text extraction across these various fonts, we aimed to determine which font type yields the best results in terms of recognition accuracy. The results, as illustrated in Table 2, show that the Battambang and Khmer fonts achieved the lowest CER and WER with Pytesseract. Battambang had a CER of approximately 0.01 and a WER of 0.07, while Khmer had a CER of 0.01 and a WER of 0.08, outperforming other fonts in terms of accuracy. The graphic displayed in Figure 2, demonstrates that the easiest Khmer fonts for Pytesseract to recognize are Khmer, Odor MeanChey, Siemreap, Sithi Manuss, and Battambang, while the most difficult to recognize are iSeth First, Bayon, and Dangrek.


\section{Discussion}
This study provides valuable insights into how different Khmer font types impact text recognition accuracy using Pytesseract. The results show significant variation in performance across fonts, with Khmer, Odor MeanChey, Siemreap, Sithi Manuss, and Battambang achieving higher accuracy due to their predictable character shapes. In contrast, iSeth First, Bayon, and Dangrek present challenges, with certain characters frequently misrecognized due to subtle visual similarities and font-specific complexities. Table 3, highlights the most difficult characters to recognize, which are found in the hardest fonts: Bayon, Dangrek, and iSeth First.

These findings highlight the need for OCR systems to improve their handling of scripts with complex typography, such as Khmer. In future work, we will explore font-specific recognition models or advanced machine learning techniques that can better adapt to the typographic diversity of Khmer script, reducing character error rates and improving recognition accuracy.

\section{Conclusion}
This research has effectively utilized the robust capabilities of Pytesseract to investigate how different font types impact text recognition accuracy. By generating a diverse dataset featuring various Khmer fonts—including Angkor, Battambang, Bayon, Bokor, Chenla, Dangrek, Freehand, Kh Kompong Chhnang, Kh SN Kampongsom, Khmer, Khmer CN Stueng Songke, Khmer Savuth Pen, Metal, Moul, Odor MeanChey, Preah Vihear, Siemreap, Sithi Manuss, and iSeth First—our study has provided valuable insights into the challenges associated with each font. The research suggests that Pytesseract performs well with fonts such as Khmer, Odor MeanChey, Siemreap, Sithi Manuss, and Battambang font, while it encounters difficulties with iSeth First, Bayon, and Dangrek font.This analysis not only reveals the strengths and limitations of Pytesseract in handling Khmer scripts but also provides insights for enhancing OCR systems to better address font-specific challenges.

\bibliographystyle{unsrt}
\bibliography{ACET_Khmer_Text_Recognition_on_Different_font}

\end{document}